\documentclass{article}
\usepackage{arxiv}
\usepackage{tcolorbox}
\usepackage{booktabs}
\usepackage{amsthm}
\usepackage{xcolor} % For coloring and color boxes
\usepackage{framed} % For framed environments
\usepackage{amsmath}
\usepackage{enumitem}
\usepackage{scrextend}
\usepackage[utf8]{inputenc} % allow utf-8 input
\usepackage[T1]{fontenc}    % use 8-bit T1 fonts
\usepackage{hyperref}       % hyperlinks
\usepackage{url}            % simple URL typesetting
\usepackage{booktabs}       % professional-quality tables
\usepackage{amsfonts}       % blackboard math symbols
\usepackage{nicefrac}       % compact symbols for 1/2, etc.
\usepackage{microtype}      % microtypography
\usepackage{lipsum}
\usepackage{graphicx}
\graphicspath{ {./images/} }
\usepackage{amsmath}
\usepackage{xcolor}
\usepackage{tgbonum}
\usepackage{titlesec}
\usepackage{fancybox}
\usepackage{tabularx}
\usepackage{algpseudocode}
\usepackage{geometry}  % Set margins for the page
\usepackage[utf8]{inputenc}  % Allows UTF-8 input
\usepackage[T1]{fontenc}  % Ensures proper encoding for accented characters
\usepackage[ruled,vlined]{algorithm2e}
\usepackage{multirow}
\usepackage{colortbl}
\usepackage{makecell}

\newcounter{subsubsubsection}[subsubsection]
\renewcommand\thesubsubsubsection{\thesubsubsection.\arabic{subsubsubsection}}
\titleformat{\subsubsubsection}
  {\normalfont\normalsize\bfseries}{\thesubsubsubsection}{1em}{}
\titlespacing*{\subsubsubsection}
{0pt}{3.25ex plus 1ex minus .2ex}{1.5ex plus .2ex}

\makeatletter
\renewcommand\paragraph{\@startsection{paragraph}{5}{\z@}%
  {3.25ex \@plus1ex \@minus.2ex}%
  {-1em}%
  {\normalfont\normalsize\bfseries}}
\renewcommand\subparagraph{\@startsection{subparagraph}{6}{\parindent}%
  {3.25ex \@plus1ex \@minus .2ex}%
  {-1em}%
  {\normalfont\normalsize\bfseries}}
\def\toclevel@subsubsubsection{4}
\def\toclevel@paragraph{5}
\def\toclevel@subparagraph{6}
\def\l@subsubsubsection{\@dottedtocline{4}{7em}{4em}}
\def\l@paragraph{\@dottedtocline{5}{10em}{5em}}
\def\l@subparagraph{\@dottedtocline{6}{14em}{6em}}
\makeatother

\titleformat*{\section}{\LARGE\bfseries}
\titleformat*{\subsection}{\Large\bfseries}
\titleformat*{\subsubsection}{\large\bfseries}
\titleformat*{\paragraph}{\large\bfseries}

\title{First Train to Generate, then Generate to Train: UnitedSynT5 for Few-Shot NLI}

\author{
 Sourav Banerjee* \\
  DataLabs\\
  United We Care\\
  \texttt{sb@unitedwecare.com} \\
   \And
  Anush Mahajan \\
  DataLabs\\
  United We Care\\
  \texttt{anush@unitedwecare.com}
  \And
Ayushi Agarwal\\
  DataLabs\\
  United We Care\\
  \texttt{ayushi@unitedwecare.com}
  \And
 Eishkaran Singh \\
  DataLabs\\
  United We Care\\
  \texttt{eishkaran@unitedwecare.com}
}

\begin{document}
\maketitle
\begin{abstract}
Natural Language Inference (NLI) tasks require identifying the relationship between sentence pairs, typically classified as entailment, contradiction, or neutrality. While the current state-of-the-art (SOTA) model, Entailment Few-Shot Learning (EFL), achieves a 93.1\% accuracy on the Stanford Natural Language Inference (SNLI) dataset, further advancements are constrained by the dataset's limitations. To address this, we propose a novel approach leveraging synthetic data augmentation to enhance dataset diversity and complexity. We present \textbf{UnitedSynT5}, an advanced extension of EFL that leverages a T5-based generator to synthesize additional premise-hypothesis pairs, which are rigorously cleaned and integrated into the training data. These augmented examples are processed within the EFL framework, embedding labels directly into hypotheses for consistency. We train a GTR-T5-XL model on this expanded dataset, achieving a new benchmark of \textbf{94.7\%} accuracy on the SNLI dataset, \textbf{94.0\%} accuracy on the E-SNLI dataset and \textbf{92.6\%} accuracy on the MultiNLI dataset, surpassing the previous SOTA models. This research demonstrates the potential of synthetic data augmentation in improving NLI models, offering a path forward for further advancements in natural language understanding tasks.
\end{abstract}


\begin{table}[htbp]
\centering
\begin{tabular}{lcc}
\hline
\textbf{Model} & \textbf{Year} & \textbf{Accuracy (\%)} \\
\hline
LSTM \cite{bowman-etal-2015-large} & 2015 & 77.6 \\
Decomposable Attention \cite{parikh2016decomposable} & 2016 & 86.3 \\
ESIM \cite{chen2017enhanced} & 2017 & 88.0 \\
GPT \cite{radford2018improving} & 2018 & 89.9 \\
BERT-large \cite{devlin2019bertpretrainingdeepbidirectional} & 2019 & 90.1 \\
XLNet-large \cite{yang2019xlnet} & 2019 & 91.6 \\
RoBERTa-large \cite{liu2019roberta} & 2019 & 91.7 \\
ALBERT-xxlarge \cite{lan2020albertlitebertselfsupervised} & 2020 & 91.8 \\
DeBERTa-v3-large \cite{he2021deberta} & 2021 & 91.9 \\
EFL \cite{yin2021entailment} & 2021 & 93.1 \\
% Flan-T5-XXL \cite{chung2022scaling} & 2022 & 93.6 \\
UnitedSynT5[3B] & 2024 & \textbf{94.7} \\
\hline
\end{tabular}
\vspace{2mm}
\caption{Chronological Progression of Accuracy on SNLI Leaderboard \cite{paperswithcodePapersWith} }
\label{tab:snli-accuracy}
\end{table}
\vspace{2mm}

\section{Introduction}
Natural Language Inference (NLI) is a fundamental task in natural language processing (NLP) that involves determining the logical relationship between two sentences: a premise and a hypothesis. These relationships are typically classified as entailment, contradiction, or neutral \cite{bowman-etal-2015-large}. NLI serves as a benchmark for evaluating machine understanding and has significant implications for various downstream applications, including question answering, text summarization, and information retrieval \cite{maccartney-manning-2009-extended}.

The evolution $^{\ref{tab:snli-accuracy}}$ of NLI models has been marked by steady improvements, each new approach addressing specific limitations of its predecessors. In 2015, the introduction of the Stanford Natural Language Inference (SNLI) dataset provided a large-scale benchmark for NLI tasks. The initial LSTM-based model achieved a modest accuracy of 77.6\% \cite{bowman-etal-2015-large}, setting the baseline for future improvements. These early models relied heavily on hand-crafted features and shallow neural networks, often utilizing lexical overlap and syntactic similarity measures to determine the relationship between sentences \cite{maccartney2008natural}. However, they struggled with complex linguistic phenomena such as negation, quantification, and world knowledge.

A significant leap came in 2016 with the Decomposable Attention model \cite{parikh2016decomposable}, which improved accuracy to 86.3\%. This model introduced attention mechanisms, allowing it to focus on relevant parts of the input sentences. The following year, the Enhanced LSTM (ESIM) model \cite{chen2017enhanced} further pushed the boundary to 88.0\%, demonstrating the potential of advanced neural architectures.

The advent of large language models (LLMs) and transfer learning techniques led to substantial performance gains. This paradigm shift was initiated by the introduction of contextual word embeddings, which capture the multifaceted meanings of words based on their surrounding context \cite{peters2018deep}. In 2018, the GPT model \cite{radford2018improving} achieved 89.9\% accuracy, showcasing the power of generative pre-training.

BERT \cite{devlin2019bertpretrainingdeepbidirectional} marked a significant milestone in 2019, achieving an accuracy of 90.1\% on SNLI. BERT's bidirectional transformer architecture and novel pre-training objectives (masked language modeling and next sentence prediction) allowed it to capture complex linguistic patterns and dependencies. The model's success demonstrated the effectiveness of unsupervised pre-training on large corpora followed by task-specific fine-tuning.

Subsequent models built upon this foundation, focusing on refining pre-training strategies and model architectures. In the same year, XLNet-large \cite{yang2019xlnet} and RoBERTa-large \cite{liu2019roberta} reached 91.6\% and 91.7\% accuracy respectively. RoBERTa optimized BERT's training process, including longer training with larger batches and dynamic masking. In 2020, ALBERT-xxlarge \cite{lan2020albertlitebertselfsupervised} slightly improved the performance to 91.8\%.

DeBERTa-v3-large \cite{he2021deberta} further pushed the boundary to 91.9\% in 2021 by introducing disentangled attention mechanisms and enhanced mask decoder architectures. These advancements highlighted the importance of precise parameter tuning and architectural innovations in improving NLI performance.

The introduction of few-shot learning paradigms represented a shift in approach, addressing the challenge of limited labeled data in specific domains\cite{banerjee2024llmshallucinateneedlive}. The Entailment Few-Shot Learning (EFL) model \cite{yin2021entailment} reformulated NLI as a binary decision problem, embedding labels directly into hypotheses. This method achieved a significant jump to 93.1\% accuracy on SNLI, demonstrating the potential of leveraging pre-trained knowledge for NLI tasks.

% Recent advancements have pushed the boundaries further by exploring multi-task learning and larger model scales. In 2022, the Flan-T5-XXL model \cite{chung2022scaling}, combined with a classification head, achieved 93.6\% accuracy on SNLI. This approach utilized instruction tuning, where the model is trained on a diverse set of NLP tasks framed as text-to-text problems.

Despite advances, challenges in NLI have persisted across models. LSTM-based approaches exhibited three limitations: (1) reliance on lexical overlap for prediction, failing on cases requiring semantic understanding, (2) inability to handle negation and quantification, and (3) difficulty in capturing dependencies between premise and hypothesis pairs. While attention mechanisms through models like Decomposable Attention and ESIM improved alignment between input sentences, these architectures struggled with examples requiring world knowledge or reasoning chains.

The emergence of language models and contextual embeddings marked a shift in NLI performance. However, these models faced challenges. Pretrained transformers like BERT and its variants showed improved but limited ability to generalize to patterns outside their training distribution. The few-shot learning approach EFL, while addressing data scarcity through label embedding, remained constrained by the limitations of static datasets.
This pattern reveals an insight: architectural changes cannot fully address NLI's challenges. The SNLI dataset, with 570,000 human-annotated premise-hypothesis pairs, captures a fraction of the phenomena and reasoning patterns in real-world scenarios. This limitation suggests that advancing NLI requires both new architectures and training data that reflects the complexity of language understanding. \cite{williams-etal-2018-broad}. 

To address the limitations in the previous architectures, we propose UnitedSynT5: a novel approach combining synthetic data augmentation with the EFL framework. Our method utilizes a T5 generator \cite{raffel2023exploringlimitstransferlearning} to create additional premise-hypothesis pairs, enhancing dataset diversity while maintaining semantic consistency. The augmented dataset is then processed through a rigorous filtering mechanism and formatted within the EFL framework. We train a GTR-T5-XL model \cite{ni2021largedualencodersgeneralizable} on this enriched dataset, leveraging the additional examples to improve generalization capacity. This approach achieves a new state-of-the-art accuracy of \textbf{94.7\%} on the SNLI dataset, \textbf{94.0\%} on E-SNLI and \textbf{92.6\%} on the MultiNLI dataset, surpassing previous benchmarks.

\section{Background}
\subsection{Natural Language Inference}

Natural Language Inference (NLI), also referred to as Recognizing Textual Entailment (RTE), is a task in natural language processing (NLP) aimed at determining the logical relationship between two sentences: a premise and a hypothesis \cite{dagan2005pascal}. The task involves integrating natural language with external knowledge, including commonsense reasoning \cite{rudinger2018knowledgeable}, to classify whether the premise entails, contradicts, or is neutral with respect to the hypothesis. Given a premise $p$ and a hypothesis $h$, the relationship $r(p,h)$ is defined as follows:

\begin{equation}
r(p,h) = 
\begin{cases} 
\text{entailment}, & \text{if } p \text{ logically implies } h, \\
\text{contradiction}, & \text{if } p \text{ logically implies } -h, \\
\text{neutral}, & \text{if neither entailment nor contradiction holds.} 
\end{cases}
\end{equation}

NLI requires understanding of semantics, pragmatics, and world knowledge, which are essential for semantic representation and inference \cite{bowman-etal-2015-large}. The complexity of NLI is influenced by factors such as semantic variability, world knowledge, and logical reasoning. 

Benchmark datasets, such as the Stanford Natural Language Inference (SNLI) corpus \cite{bowman-etal-2015-large},MultiNLI \cite{williams-etal-2018-broad}, cross-lingual NLI for multiple languages \cite{conneau2018xnli}, and  provide large-scale collections of human-annotated sentence pairs, designed to facilitate the evaluation and training of NLI systems. The SNLI corpus consists of 570,000 premise-hypothesis pairs, each labeled as entailment, contradiction, or neutral. These examples were generated through crowdsourcing, ensuring a diverse range of sentence structures and linguistic phenomena, making SNLI a foundational benchmark for NLI tasks. Similarly, the MultiNLI dataset extends SNLI by offering premise-hypothesis pairs from multiple genres, thus enabling the evaluation of models in a broader array of linguistic contexts and improving their ability to generalize across domains.

Early approaches of NLI primarily relied on hand-crafted features and statistical models\cite{maccartney-manning-2009-extended}. These methods involved manually designed features, such as lexical overlap, syntactic similarity, and heuristic rules, to approximate the relationship between the premise and the hypothesis. Statistical classifiers, such as Support Vector Machines (SVMs) and logistic regression, were then applied to these features to predict entailment relations. However, these approaches had limited capacity to handle more complex semantic phenomena, such as negation, modality, or world knowledge, and were highly dependent on the quality and coverage of the predefined features. 

Advancements in deep learning and transformer architectures \cite{vaswani2017attention} have led to significant improvements. Pre-trained models, such as BERT \cite{devlin2019bertpretrainingdeepbidirectional} and RoBERTa \cite{liu2019roberta}, have demonstrated superior performance when fine-tuned on NLI tasks by leveraging their large-scale pre-training.

Challenges in NLI include out-of-domain generalization, handling complex reasoning scenarios, and addressing dataset biases. These areas continue to be the focus of ongoing research aimed at improving the robustness and generalization of NLI models.

\subsection{Entailment Few-Shot Learning (EFL)}
The Entailment Few-Shot Learning (EFL) approach \cite{yin2021entailment} demonstrates state-of-the-art performance on the SNLI dataset, achieving an accuracy of 93.1\%. EFL reformulates the NLI task by embedding the label directly into the hypothesis, transforming the traditional three-way classification into a binary decision problem. This method allows the model to focus on verifying the relationship between the premise and the label associated with the hypothesis. While this reformulation has proven effective, it may overlook the more subtle and complex relationships that arise in certain NLI tasks.

For example, in the EFL framework, a pair of sentences is modified as follows:

\textit{\textbf{Premise}: A yellow bulldozer moves dirt in a rocky field.} \\ 
\textit{\textbf{Hypothesis}: It is raining.}

In this instance, the relationship between the premise and hypothesis is labeled as \textit{neutral}. EFL reformulates the hypothesis by embedding the label, resulting in:

\textit{\textbf{Premise}: A yellow bulldozer moves dirt in a rocky field.\\ \textbf{Hypothesis}: It is raining. This is neutral.}

The model’s task is then to determine whether the extended hypothesis is "true" or "false," simplifying the classification into a binary decision.

This approach provides a streamlined method for assessing entailment but may also introduce limitations when handling more multifaceted linguistic relationships.

\section{Methodology: Overcoming the Current Performance Limitations}

\begin{figure*}
    \centering
    \includegraphics[width=\linewidth]{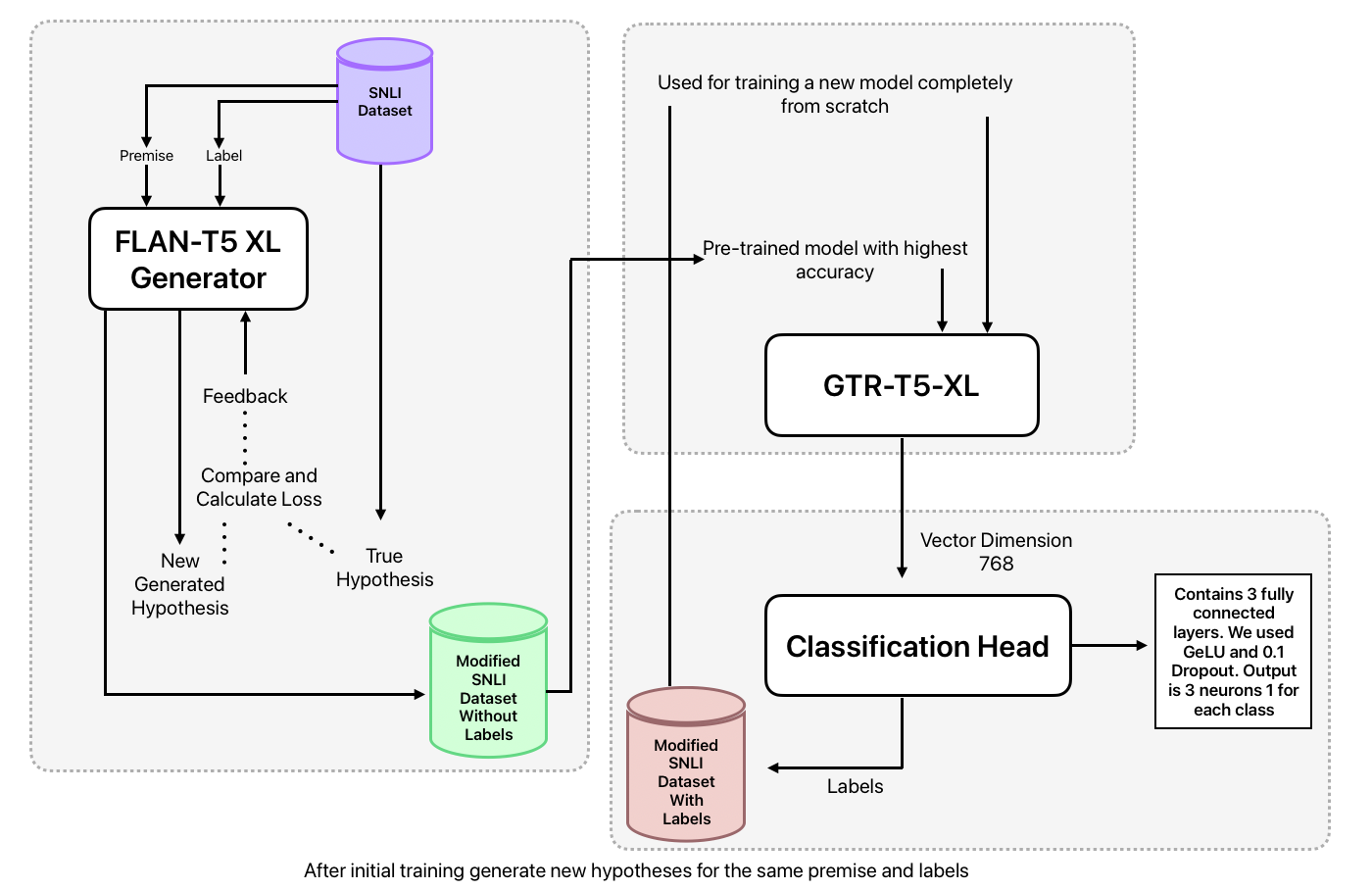}
    \caption{Overview of the proposed approach for enhancing Natural Language Inference (NLI) performance through synthetic data augmentation.}
    \label{fig:Process-Diagram}
\end{figure*}

We hypothesize that the current performance ceiling in Natural Language Inference (NLI) tasks, such as those measured on the SNLI dataset, is due to the limited variation and complexity within the dataset. To address this, we propose expanding the dataset through synthetic data generation, aiming to introduce greater linguistic diversity and complexity.$^{\ref{fig:Process-Diagram},\ref{Algo:Algorithm-1}}$

\subsection{Generation of Synthetic Dataset}
The generation of a synthetic dataset \cite{banerjee2024highprecisionmedicalspeechrecognition} uses an extension of the Entailment Few-Shot Learning (EFL) model. The process leverages a FLAN-T5 XL (3B) generator to produce additional premise-hypothesis pairs without requiring manual labeling. The following steps outline the structure and workflow involved in generating this dataset:
\begin{enumerate} 
    \item The initial training dataset is partitioned into two subsets: a generation set and a few-shot example set. This partitioning is based on a predefined 95\%-5\% split, where 95\% of the data (521,898 examples) is allocated for generation and 5\% (27,469 examples) for few-shot learning.
    \item For each example within the generation set, two examples from the few-shot set are randomly selected and included as contextual references in the generation prompt. These examples guide the generator during hypothesis generation, providing additional context to enhance the generation process. 
\end{enumerate}
The FLAN-T5 XL generator is employed to produce new hypotheses based on the premises found in the generation set. By incorporating the two few-shot examples into the prompt, the generator improves its understanding of the NLI task, allowing for more contextually appropriate hypothesis generation, ensuring that the generated hypotheses are semantically consistent with the premises and reflect meaningful entailment relationships.

The generated synthetic dataset is then utilized to train the GTR-T5-XL model, which performs classification tasks based on the augmented data. The iterative training and data generation process, visualized in the approach diagram, ensures that the model benefits from the increased diversity of the augmented dataset, leading to improved generalization capabilities on classification tasks.
\vspace{-2.5mm}
\begin{algorithm}[H]
\SetAlgoLined
\KwIn{Original SNLI dataset $D_{SNLI}$}
\KwOut{Trained GTR-T5-XL model $M_{GTR}$}

\textbf{Step 1: Generate Synthetic Dataset}
$D_{gen}, D_{few} \gets \text{SplitDataset}(D_{SNLI}, 0.95, 0.05)$\; 
Initialize FLAN-T5 XL generator $G$\; 
$D_{syn} \gets \emptyset$\;
\For{each premise $P$ in $D_{gen}$}{    $E_1, E_2 \gets \text{RandomSample}(D_{few}, 2)$\;
    $prompt \gets \text{ConstructPrompt}(P, E_1, E_2)$\;
    $H \gets G(prompt)$\;
    $D_{syn} \gets D_{syn} \cup \{(P, H)\}$\;
}

\textbf{Step 2: Clean Dataset}
$D_{clean} \gets \emptyset$\;
\For{each $(P, H, L)$ in $D_{syn}$}{
    $L_{SOTA} \gets \text{PredictSOTA}(P, H)$\;
    \If{$L = L_{SOTA}$ and $(P, H) \notin D_{SNLI}$}{
        $D_{clean} \gets D_{clean} \cup \{(P, H, L)\}$\;
    }
}

\textbf{Step 3: EFL Conversion}
$D_{EFL} \gets \emptyset$\;
\For{each $(P, H, L)$ in $D_{clean}$}{
    $H_{EFL} \gets \text{"The hypothesis '} + H + \text{' is a } + L + \text{ of the premise."}$\;
    $D_{EFL} \gets D_{EFL} \cup \{(P, H_{EFL})\}$\;
}

\textbf{Step 4: Train UnitedSynT5 Model}
Initialize UnitedSynT5L model $M_{GTR}$\;
Configure $M_{GTR}$ with input dimension 768, 3 FC layers, GeLU activation, dropout 0.1\;
\While{not converged}{
    \For{batch $b$ in $D_{EFL}$ and human-labeled SNLI dataset}
    {
        $loss \gets \text{ComputeLoss}(M_{GTR}(b), \text{TrueLabels}(b))$\;
        Update $M_{GTR}$ parameters using gradient of $loss$\;
    }
}

\textbf{Step 5: Evaluate Model}
$performance \gets \text{EvaluateModel}(M_{GTR}, D_{SNLI})$\;

\Return $M_{GTR}$\\
\caption{Augmented Entailment: Enhancing Few-Shot Natural
Language Inference with Synthetic Data Generation}
\label{Algo:Algorithm-1}
\end{algorithm}

\subsection{Training the FLAN-T5 XL Generator to Generate Hypotheses}

The training process for the FLAN-T5 XL generator involved the application of a few-shot learning approach. In this setup, two examples were randomly selected from 5\% of the total training data to serve as additional context for the generator during hypothesis generation. This few-shot learning approach was adopted to address both computational and contextual limitations of the model, optimizing the balance between the input data size and model performance without the need for a large-scale dataset.

The few-shot examples served as guiding references during training, allowing the model to maintain coherence with the premise and align with the task's entailment, contradiction, or neutrality labels.

\subsubsection{Prompt Design and Structure}
The prompts used in the training process were designed to be highly structured, ensuring that the generated hypotheses adhered to the standards of the Stanford Natural Language Inference (SNLI) task. Each prompt explicitly defined the task, labels, and guidelines to maintain consistency across different training examples. Below is an outline of the prompt structure used during the generation process:

\begin{tcolorbox}[colback=white, colframe=black, width=\textwidth, arc=0mm, auto outer arc]
\textbf{Task}: Generate a concise hypothesis corresponding to the given premise and label, in accordance with the SNLI task.

\textbf{Labels}: 'entailment', 'contradiction', 'neutral'.

\textbf{Guidelines:}
\begin{itemize}
    \item \textbf{Entailment}: The hypothesis logically follows from the premise.
    \item \textbf{Contradiction}: The hypothesis contradicts the premise.
    \item \textbf{Neutral}: The hypothesis is unrelated to the premise.
\end{itemize}

\textbf{Examples:}
\begin{itemize}
    \item \textbf{Label: Entailment}
    \begin{itemize}
        \item \textbf{Premise}: A person on a horse jumps over a broken-down airplane.
        \item \textbf{Hypothesis}: A person is outdoors, on a horse.
    \end{itemize}
    
    \item \textbf{Label: Contradiction}
    \begin{itemize}
        \item \textbf{Premise}: Children are smiling and waving at the camera.
        \item \textbf{Hypothesis}: The kids are frowning.
    \end{itemize}
\end{itemize}

\textbf{Your Task:}
\begin{itemize}
    \item \textbf{Label: Contradiction}
    \begin{itemize}
        \item \textbf{Premise}: Two blond women are hugging one another.
        \item \textbf{Hypothesis}: The women are arguing with each other.
    \end{itemize}
\end{itemize}
\end{tcolorbox}

The inclusion of detailed task instructions and examples ensured that the generator produced hypotheses with high alignment to the SNLI dataset's requirements. The structured prompts guided the generator in producing hypotheses that were semantically and contextually consistent with the corresponding premises and labels.
\subsubsection{Iterative Process for Generator Training}
The training process for the FLAN-T5 XL generator employed an iterative approach, wherein the generated hypotheses were evaluated against the reference hypotheses from the SNLI dataset. During each iteration, the generator produced a set of candidate hypotheses based on the input premises, which were then compared with the true labels (entailment, contradiction, or neutral) provided by the dataset. To measure the discrepancy between the predicted and actual labels, a cross-entropy loss function was applied. This function is commonly used in classification tasks to evaluate the difference between the predicted probability distribution and the actual label, providing a penalty for incorrect predictions based on their assigned probability.

The calculated loss was then backpropagated through the model, enabling parameter updates using a gradient descent optimization algorithm, such as Adam. The model's parameters, including the weights within the neural network, were adjusted to minimize the loss in subsequent iterations, thereby improving its performance in generating hypotheses that aligned with the true labels. This iterative feedback loop, driven by loss minimization, enhanced the model's ability to generalize across different premise-hypothesis pairs.

The model's parameters were adjusted in each iteration based on the calculated loss, improving the generator's capacity to produce hypotheses that aligned with the true labels. The iterative nature of the training process ensured that the generator's performance gradually improved, resulting in a more robust model capable of producing accurate hypotheses across different categories.

\subsubsection{Output and Dataset Augmentation}
Upon completion of the training process, the generator was tasked with producing a set of synthetic examples, consisting of premise-hypothesis pairs, based on the prompts and training data. These generated examples were then integrated into the training dataset, augmenting the original SNLI data with new, machine-generated hypotheses.

\subsection{Dataset Cleaning}

To maintain the integrity of the augmented dataset, a systematic filtering process was applied to remove inconsistent or redundant data. This process was designed to ensure that the synthetic examples generated during the data augmentation stage adhered to the expected quality standards, thus preventing any negative impact on model training.

\subsubsection{Filtering Criteria}

The filtering process was based on two primary criteria, each aimed at refining the dataset to preserve both accuracy and diversity:

\begin{enumerate}
    \item \textbf{Label Alignment}: The first criterion ensured that the labels generated by the FLAN-T5 XL generator matched the labels assigned by the GTR-T5-XL model, which is the current state-of-the-art (SOTA) model for the task. Any instance where the generated label did not align with the GTR-T5-XL model’s prediction was removed. This helped in maintaining consistency between the generated data and the labels used for training, thereby preventing the inclusion of data that could degrade the model's performance.
    
    \item \textbf{Redundancy Elimination}: The second criterion focused on removing any hypotheses that exactly replicated examples from the existing training data. Retaining such duplicates would not contribute to the dataset's diversity and could introduce redundancy. By eliminating these duplicates, the augmented dataset retained only novel examples, thus enhancing its utility in training the model.
\end{enumerate}

\subsubsection{Dataset Reduction}
The filtering process led to the removal of 54,216 examples from the initial set of 521,899 generated examples, leaving 467,683 examples in the final cleaned dataset. The refined dataset formed a more robust foundation for training, minimizing the risks associated with overfitting or noise due to redundant or mislabeled data.

Below are examples of retained generated hypotheses, demonstrating the application of the filtering criteria and the resulting dataset quality:

\begin{tcolorbox}[colback=white, colframe=black, width=\textwidth, arc=0mm, auto outer arc]
\begin{itemize}
    \item \textbf{Premise}: A young blond boy is eating a banana while an elderly woman in the background watches.\newline
    \textbf{Label}: Contradiction\newline
    \textbf{Generated Hypothesis}: A young boy is eating a kiwi.
    
    \item \textbf{Premise}: Several people paying attention to a lecture.\newline
    \textbf{Label}: Neutral\newline
    \textbf{Generated Hypothesis}: The people are in a classroom.
    
    \item \textbf{Premise}: A man on patio furniture drinking a beer. \newline
    \textbf{Label}: Contradiction \newline
    \textbf{Generated Hypothesis}: A man is drinking a soda.
\end{itemize}
\end{tcolorbox}

The data reduction process maintained a balance between dataset size and quality. By focusing on label alignment and redundancy elimination, the process ensured that the augmented dataset remained both diverse and accurate, enhancing the robustness of the model. 

The filtered dataset was subsequently used to train the GTR-T5-XL model, ensuring exposure to high-quality, diverse examples aligned with the task requirements. This enabled the model to learn from a wider variety of premise-hypothesis pairs, contributing to improved and generalizable performance on natural language inference tasks.

\subsection{EFL Conversion}

The conversion of the synthetic dataset to the Entailment Few-Shot Learning (EFL) framework constitutes a critical step in the methodology. This process involves the systematic reformulation of the generated data to align with the EFL paradigm, which integrates label information directly into the hypothesis statements. The conversion facilitates the incorporation of synthetic examples into existing EFL-based training pipelines and ensures methodological consistency.

In the EFL framework, the task undergoes a transformation from a three-way classification problem to a binary decision task which is through the embedding of entailment labels within the hypothesis itself. The model is then tasked with determining the veracity of the entire statement, including the embedded label, as either "true" or "false". 

The EFL conversion process can be formalized as follows:

Let $P$ be the premise, $H$ the original hypothesis, and $L$ the label (entailment, contradiction, or neutral). The EFL conversion function $f_{EFL}$ can be defined as:

\[
f_{EFL}(P, H, L) = (P, H_{EFL})
\]

Where $H_{EFL}$ is the modified hypothesis incorporating the label information. For example:

\[
H_{EFL} = \text{"The hypothesis '} + H + \text{' is a} + L + \text{ of the premise."}
\]

This conversion has several implications:

\begin{itemize}
    \item \textbf{Data Augmentation}: The EFL conversion expands the training corpus by creating multiple variations of each premise-hypothesis pair. Each original example yields three EFL-formatted instances, one for each possible label. This augmentation can be represented as:
    
    \[
    |D_{EFL}| = 3 \times |D_{original}|
    \]
    
    Where $D_{EFL}$ is the EFL-converted dataset and $D_{original}$ is the original dataset.

    \item \textbf{Semantic Relationship Enhancement}: By explicitly verbalizing the label within the hypothesis, the model is compelled to evaluate the semantic connections between the premise, hypothesis, and the purported relationship. This can be conceptualized as a function $g$ that maps the input to a semantic space $S$:
    
    \[
    g: (P, H_{EFL}) \rightarrow S
    \]
    
    The model learns to optimize this mapping to accurately discern the true relationships in the semantic space.
\end{itemize}

The application of the EFL framework to the synthetic dataset ensures interoperability between human-annotated and machine-generated examples. This integration maintains training consistency while introducing increased linguistic variability into the dataset. The process can be described algorithmically:

\begin{algorithm}[H]
\SetAlgoLined
\KwIn{Synthetic dataset $D_{syn} = \{(P_i, H_i, L_i)\}_{i=1}^N$}
\KwOut{EFL-converted dataset $D_{EFL}$}
\SetKwFunction{ConstructEFLHypothesis}{ConstructEFLHypothesis}
Initialize $D_{EFL} \gets \emptyset$\;
\For{$i \gets 1$ \KwTo $N$}{
    $P \gets P_i$\;
    $H \gets H_i$\;
    $L \gets L_i$\;
    $H_{EFL} \gets \ConstructEFLHypothesis(H, L)$\;
    $D_{EFL} \gets D_{EFL} \cup \{(P, H_{EFL})\}$\;
}
\caption{EFL Conversion of Synthetic Dataset}
\end{algorithm}

Where \ConstructEFLHypothesis is a function that generates the EFL-formatted hypothesis as described earlier. This function can be defined as:

\begin{algorithm}[H]
\SetAlgoLined
\KwIn{Original hypothesis $H$, Label $L$}
\KwOut{EFL-formatted hypothesis $H_{EFL}$}
\SetKwFunction{ConstructEFLHypothesis}{ConstructEFLHypothesis}
\ConstructEFLHypothesis{$H$, $L$}:\;
\Begin{
    $H_{EFL} \gets \text{"The hypothesis `"} + H + \text{"' is a "} + L + \text{" of the premise."}$\;
    \KwRet $H_{EFL}$\;
}
\caption{ConstructEFLHypothesis Function}
\end{algorithm}

The EFL conversion process interacts with the subsequent training phase by modifying the input format and the learning objective. Instead of training on $(P, H, L)$ triples, the model now trains on $(P, H_{EFL})$ pairs with a binary classification objective. 

\subsection{Training a New GTR-T5-XL Model}

The process of training the GTR-T5-XL begins with the use of the synthetic dataset generated in the previous steps. This dataset, which has been cleaned and processed through the Entailment Few-Shot Learning (EFL) framework, was utilized to train the GTR-T5-XL model from the ground up.

The GTR-T5-XL model was initially pre-trained on a large corpus, providing it with a broad understanding of natural language. Subsequently, it was fine-tuned using the newly augmented synthetic dataset along with human labeled training set either produced by the FLAN-T5 XL generator or sourced from the original human-labeled data. This dataset, containing machine-generated hypotheses for each premise, was constructed to enhance the model’s capability in solving the Natural Language Inference (NLI) task. 

\subsubsection{Training Configuration}

The training process involved providing the GTR-T5-XL model with input embeddings, each of a vector dimension of 768. The model included three fully connected layers. These layers were configured with the Gaussian Error Linear Unit (GeLU) activation function, which was employed to introduce non-linearity and improve the model’s ability to capture complex relationships in the data. Additionally, a dropout rate of 0.1 was used throughout the network to prevent overfitting during the training process.

The final classification layer consisted of three output neurons, each representing one of the NLI classification categories: entailment, contradiction, and neutral. This configuration allowed the model to predict the appropriate label for each premise-hypothesis pair based on the learned representations from both human-labeled and machine-generated data.

\subsubsection{Evaluation and Performance Measurement}

After the completion of training, the model was evaluated on the original SNLI dataset to assess its performance. The evaluation process involved comparing the model's predictions with the human-annotated labels within the dataset. This validation provided a direct measurement of the effectiveness of the synthetic dataset in improving the model's ability to perform NLI tasks. The use of human-annotated labels as a benchmark enabled a reliable comparison to determine the extent to which the augmented data contributed to the model’s improved performance.

\subsubsection{Feedback and Model Refinement}

The training process incorporated a feedback loop to continuously refine the model's parameters. This iterative refinement allowed the GTR-T5-XL model to generalize more effectively when exposed to unseen data. By adjusting the model’s parameters based on performance during training, the feedback loop ensured that the model was consistently improving its accuracy and ability to handle complex and diverse premise-hypothesis pairs.

\section{Results}

The experimental results, demonstrating how our approach surpasses the current state-of-the-art (SOTA) in terms of test accuracy. We employed the FLAN-T5 XL model (3 billion parameters) for generation tasks, paired with GTR-Large models for classification tasks of varying sizes. The integration of our augmented dataset yielded improvements over the previous SOTA results.

\begin{table}[htbp]
\centering
\renewcommand{\arraystretch}{1.2}
\begin{tabular}{|l|c|l|l|l|}
\hline
\textbf{Benchmark} & \textbf{Year} & \textbf{Model} & \textbf{Parameters} & \textbf{Performance} \\
\hline
\multirow{12}{*}{SNLI} 
& 2015 & LSTM \cite{bowman-etal-2015-large} & - & 77.6\% \\
& 2016 & Decomposable Attention \cite{parikh2016decomposable} & - & 86.3\% \\
& 2017 & ESIM \cite{chen2017enhanced} & - & 88.0\% \\
& 2018 & GPT \cite{radford2018improving} & 117M & 89.9\% \\
& 2019 & BERT-large \cite{devlin2019bertpretrainingdeepbidirectional} & 340M & 90.1\% \\
& 2019 & XLNet-large \cite{yang2019xlnet} & 340M & 91.6\% \\
& 2019 & RoBERTa-large \cite{liu2019roberta} & 355M & 91.7\% \\
& 2020 & ALBERT-xxlarge \cite{lan2020albertlitebertselfsupervised} & 223M & 91.8\% \\
& 2021 & DeBERTa-v3-large \cite{he2021deberta} & 350M & 91.9\% \\
& 2021 & EFL \cite{yin2021entailment} & - & 93.1\% \\
\rowcolor{white}&
\fcolorbox{red}{white}{2024} & 
\fcolorbox{red}{white}{UnitedSynT5 (335M)} &
\fcolorbox{red}{white}{335M} &
\fcolorbox{red}{white}{93.5\%} \\
\rowcolor{white}&
\fcolorbox{red}{white}{2024} & 
\fcolorbox{red}{white}{UnitedSynT5 (3B)} &
\fcolorbox{red}{white}{3B} &
\fcolorbox{red}{white}{94.7\%} \\
\hline
\multirow{6}{*}{E-SNLI}
& 2019 & BERT-large \cite{devlin2019bertpretrainingdeepbidirectional} & 340M & 89.5\% \\
& 2019 & RoBERTa-large \cite{liu2019roberta} & 355M & 90.7\% \\
& 2020 & BART-large \cite{lewis2019bartdenoisingsequencetosequencepretraining} & 406M & 92.3\% \\
& 2020 & T5-large \cite{raffel2023exploringlimitstransferlearning} & 770M & 91.8\% \\
\rowcolor{white}&
\fcolorbox{red}{white}{2024} & 
\fcolorbox{red}{white}{UnitedSynT5 (335M)} &
\fcolorbox{red}{white}{335M} &
\fcolorbox{red}{white}{89.8\%} \\
\rowcolor{white}&
\fcolorbox{red}{white}{2024} & 
\fcolorbox{red}{white}{UnitedSynT5 (3B)} &
\fcolorbox{red}{white}{3B} &
\fcolorbox{red}{white}{94.0\%} \\
\hline
\multirow{10}{*}{MultiNLI}
& 2018 & BERT-large \cite{devlin2019bertpretrainingdeepbidirectional} & 340M & 86.7\% \\
& 2019 & RoBERTa-large \cite{liu2019roberta} & 355M & 90.2\% \\
& 2021 & DeBERTa-v3-large \cite{he2021deberta} & 375M & 91.1\%  \\
& 2021 & Turing NLR v5 XXL \cite{Jiang_2020} & 5.4B & 92.6\% \\
& 2022 & GPT-3.5 \cite{radford2018improving} & 175B & 90.2\% \\
& 2023 & PaLM-2 \cite{anil2023palm2technicalreport} & 340B$*$ & 90.8\% \\
& 2024 & Claude 3 \cite{anthropic2023claude3} & 400B & 91.2\% \\
& 2024 & GPT-4 \cite{openai2024gpt4technicalreport} & 1.7T & 91.8\% \\
\rowcolor{white}&
\fcolorbox{red}{white}{2024} & 
\fcolorbox{red}{white}{UnitedSynT5 (335M)} &
\fcolorbox{red}{white}{335M} &
\fcolorbox{red}{white}{89.8\%} \\
\rowcolor{white}&
\fcolorbox{red}{white}{2024} & 
\fcolorbox{red}{white}{UnitedSynT5 (3B)} &
\fcolorbox{red}{white}{3B} &
\fcolorbox{red}{white}{92.6\%} \\
\hline
% \multicolumn{6}{l}{\small *Estimated parameters} 
\end{tabular}
\vspace{2mm}
\caption{Comprehensive performance comparison of NLI models across different benchmarks}
\label{tab:nli_comparison}r
\end{table}

As shown in Table$^{\ref{tab:nli_comparison}}$, the combination of Flan-T5-xl (3B) for generation with GTR-Large (335M) for classification achieved a test accuracy of 93.5\%, surpassing the previous SOTA accuracy of 93.1\%. Further gains were observed when using the larger GTR-Large (3B) model, which improved test accuracy to \textbf{94.7\%}.

In addition to the SNLI dataset, the same approach was tested on the E-SNLI and MultiNLI datasets, where it also broke previous records, setting new accuracies of \textbf{94.0\%} on E-SNLI  and \textbf{92.6\%} on MultiNLI. These results confirm the generalizability of our method across multiple NLI benchmarks, establishing new SOTA performance on each.

These results highlight the effectiveness of model scaling and synthetic data augmentation in improving NLI performance. The increased accuracy demonstrates that incorporating a well-augmented dataset enhances model generalization and performance, particularly when larger classification models are employed.

\section{Limitations and Future Work}
The current approach faces several limitations related to computational efficiency and model performance. One limitation concerns computational constraints, where the number of tokens in the input prompt increases the computational resources required for hypothesis generation exponentially. This increase in token count also leads to a significant extension in training time, reducing the efficiency of the approach in large-scale applications. Another issue is related to maintaining contextual coherence, especially in smaller models such as T5-3B. These models encounter difficulties in handling longer prompts, making it challenging to incorporate a large number of examples within the prompt while preserving context and coherence. This constraint affects the model's ability to process complex inputs effectively.

Future work could explore several potential improvements. One avenue involves testing different ratios for the train/few-shot data split and varying the number of few-shot examples included in the prompt. This experimentation may help identify configurations that optimize model performance relative to input size. Additionally, while synthetic dataset augmentation has been applied to state-of-the-art models in Natural Language Inference (NLI) tasks, further research could investigate the impact of this technique on other NLI approaches beyond the Entailment Few-Shot Learning (EFL) framework. Exploring its effectiveness across different methodologies could provide further insights into its broader applicability in NLI model development.

\bibliographystyle{ieeetr}  
\nocite{*}
\bibliography{main.bib}  %%% Remove comment to use the external .bib file (using bibtex).

\begin{thebibliography}{10}

\bibitem{bowman-etal-2015-large}
S.~R. Bowman, G.~Angeli, C.~Potts, and C.~D. Manning, ``A large annotated corpus for learning natural language inference,'' in {\em Proceedings of the 2015 Conference on Empirical Methods in Natural Language Processing} (L.~M{\`a}rquez, C.~Callison-Burch, and J.~Su, eds.), (Lisbon, Portugal), pp.~632--642, Association for Computational Linguistics, Sept. 2015.

\bibitem{parikh2016decomposable}
A.~Parikh, O.~T{\"a}ckstr{\"o}m, D.~Das, and J.~Uszkoreit, ``A decomposable attention model for natural language inference,'' in {\em Proceedings of the 2016 Conference on Empirical Methods in Natural Language Processing}, pp.~2249--2255, 2016.

\bibitem{chen2017enhanced}
Q.~Chen, X.~Zhu, Z.~Ling, S.~Wei, H.~Jiang, and D.~Inkpen, ``Enhanced lstm for natural language inference,'' in {\em Proceedings of the 55th Annual Meeting of the Association for Computational Linguistics (Volume 1: Long Papers)}, pp.~1657--1668, 2017.

\bibitem{radford2018improving}
A.~Radford, K.~Narasimhan, T.~Salimans, and I.~Sutskever, ``Improving language understanding by generative pre-training,'' 2018.
\newblock OpenAI preprint.

\bibitem{devlin2019bertpretrainingdeepbidirectional}
J.~Devlin, M.-W. Chang, K.~Lee, and K.~Toutanova, ``Bert: Pre-training of deep bidirectional transformers for language understanding,'' 2019.

\bibitem{yang2019xlnet}
Z.~Yang, Z.~Dai, Y.~Yang, J.~Carbonell, R.~R. Salakhutdinov, and Q.~V. Le, ``Xlnet: Generalized autoregressive pretraining for language understanding,'' {\em Advances in neural information processing systems}, vol.~32, 2019.

\bibitem{liu2019roberta}
Y.~Liu, M.~Ott, N.~Goyal, J.~Du, M.~Joshi, D.~Chen, O.~Levy, M.~Lewis, L.~Zettlemoyer, and V.~Stoyanov, ``Roberta: A robustly optimized bert pretraining approach,'' in {\em arXiv preprint arXiv:1907.11692}, 2019.

\bibitem{lan2020albertlitebertselfsupervised}
Z.~Lan, M.~Chen, S.~Goodman, K.~Gimpel, P.~Sharma, and R.~Soricut, ``Albert: A lite bert for self-supervised learning of language representations,'' 2020.

\bibitem{he2021deberta}
P.~He, J.~Gao, W.-t. Yih, and X.~Deng, ``Deberta: Decoding-enhanced bert with disentangled attention,'' {\em arXiv preprint arXiv:2106.03654}, 2021.

\bibitem{yin2021entailment}
W.~Yin, J.~Hay, T.~Khot, A.~Sabharwal, P.~Clark, and D.~Roth, ``Entailment as few-shot learner,'' in {\em Proceedings of the 2021 Conference on Empirical Methods in Natural Language Processing}, pp.~9368--9383, 2021.

\bibitem{paperswithcodePapersWith}
``{P}apers with {C}ode - {S}{N}{L}{I} {B}enchmark ({N}atural {L}anguage {I}nference) --- paperswithcode.com.'' \url{https://paperswithcode.com/sota/natural-language-inference-on-snli}.
\newblock [Accessed 18-10-2024].

\bibitem{maccartney-manning-2009-extended}
B.~MacCartney and C.~D. Manning, ``An extended model of natural logic,'' in {\em Proceedings of the Eight International Conference on Computational Semantics} (H.~Bunt, ed.), (Tilburg, The Netherlands), pp.~140--156, Association for Computational Linguistics, Jan. 2009.

\bibitem{maccartney2008natural}
B.~MacCartney, ``Natural language inference,'' in {\em Stanford University}, 2008.

\bibitem{peters2018deep}
M.~E. Peters, M.~Neumann, M.~Iyyer, M.~Gardner, C.~Clark, K.~Lee, and L.~Zettlemoyer, ``Deep contextualized word representations,'' in {\em Proceedings of the 2018 Conference of the North American Chapter of the Association for Computational Linguistics: Human Language Technologies, Volume 1 (Long Papers)}, pp.~2227--2237, 2018.

\bibitem{banerjee2024llmshallucinateneedlive}
S.~Banerjee, A.~Agarwal, and S.~Singla, ``Llms will always hallucinate, and we need to live with this,'' 2024.

\bibitem{williams-etal-2018-broad}
A.~Williams, N.~Nangia, and S.~Bowman, ``A broad-coverage challenge corpus for sentence understanding through inference,'' in {\em Proceedings of the 2018 Conference of the North {A}merican Chapter of the Association for Computational Linguistics: Human Language Technologies, Volume 1 (Long Papers)} (M.~Walker, H.~Ji, and A.~Stent, eds.), (New Orleans, Louisiana), pp.~1112--1122, Association for Computational Linguistics, June 2018.

\bibitem{raffel2023exploringlimitstransferlearning}
C.~Raffel, N.~Shazeer, A.~Roberts, K.~Lee, S.~Narang, M.~Matena, Y.~Zhou, W.~Li, and P.~J. Liu, ``Exploring the limits of transfer learning with a unified text-to-text transformer,'' 2023.

\bibitem{ni2021largedualencodersgeneralizable}
J.~Ni, C.~Qu, J.~Lu, Z.~Dai, G.~H. Ábrego, J.~Ma, V.~Y. Zhao, Y.~Luan, K.~B. Hall, M.-W. Chang, and Y.~Yang, ``Large dual encoders are generalizable retrievers,'' 2021.

\bibitem{dagan2005pascal}
I.~Dagan, O.~Glickman, and B.~Magnini, ``The pascal recognising textual entailment challenge,'' in {\em Machine Learning Challenges Workshop}, pp.~177--190, Springer, 2005.

\bibitem{rudinger2018knowledgeable}
R.~Rudinger, S.~Istvan, and B.~Van~Durme, ``Knowledgeable reader: Enhancing cloze-style reading comprehension with external commonsense knowledge,'' in {\em Proceedings of the 56th Annual Meeting of the Association for Computational Linguistics (Volume 1: Long Papers)}, pp.~1753--1762, 2018.

\bibitem{conneau2018xnli}
A.~Conneau, R.~Rinott, G.~Lample, A.~Williams, S.~R. Bowman, H.~Schwenk, and V.~Stoyanov, ``Xnli: Evaluating cross-lingual sentence representations,'' in {\em Proceedings of the 2018 Conference on Empirical Methods in Natural Language Processing}, pp.~2475--2485, 2018.

\bibitem{vaswani2017attention}
A.~Vaswani, N.~Shazeer, N.~Parmar, J.~Uszkoreit, L.~Jones, A.~N. Gomez, {\L}.~Kaiser, and I.~Polosukhin, ``Attention is all you need,'' in {\em Advances in neural information processing systems}, pp.~5998--6008, 2017.

\bibitem{banerjee2024highprecisionmedicalspeechrecognition}
S.~Banerjee, A.~Agarwal, and P.~Ghosh, ``High-precision medical speech recognition through synthetic data and semantic correction: United-medasr,'' 2024.

\bibitem{lewis2019bartdenoisingsequencetosequencepretraining}
M.~Lewis, Y.~Liu, N.~Goyal, M.~Ghazvininejad, A.~Mohamed, O.~Levy, V.~Stoyanov, and L.~Zettlemoyer, ``Bart: Denoising sequence-to-sequence pre-training for natural language generation, translation, and comprehension,'' 2019.

\bibitem{Jiang_2020}
H.~Jiang, P.~He, W.~Chen, X.~Liu, J.~Gao, and T.~Zhao, ``Smart: Robust and efficient fine-tuning for pre-trained natural language models through principled regularized optimization,'' in {\em Proceedings of the 58th Annual Meeting of the Association for Computational Linguistics}, Association for Computational Linguistics, 2020.

\bibitem{anil2023palm2technicalreport}
R.~Anil, A.~M. Dai, O.~Firat, M.~Johnson, D.~Lepikhin, A.~Passos, S.~Shakeri, E.~Taropa, P.~Bailey, Z.~Chen, E.~Chu, J.~H. Clark, L.~E. Shafey, Y.~Huang, K.~Meier-Hellstern, G.~Mishra, E.~Moreira, M.~Omernick, K.~Robinson, S.~Ruder, Y.~Tay, K.~Xiao, Y.~Xu, Y.~Zhang, G.~H. Abrego, J.~Ahn, J.~Austin, P.~Barham, J.~Botha, J.~Bradbury, S.~Brahma, K.~Brooks, M.~Catasta, Y.~Cheng, C.~Cherry, C.~A. Choquette-Choo, A.~Chowdhery, C.~Crepy, S.~Dave, M.~Dehghani, S.~Dev, J.~Devlin, M.~Díaz, N.~Du, E.~Dyer, V.~Feinberg, F.~Feng, V.~Fienber, M.~Freitag, X.~Garcia, S.~Gehrmann, L.~Gonzalez, G.~Gur-Ari, S.~Hand, H.~Hashemi, L.~Hou, J.~Howland, A.~Hu, J.~Hui, J.~Hurwitz, M.~Isard, A.~Ittycheriah, M.~Jagielski, W.~Jia, K.~Kenealy, M.~Krikun, S.~Kudugunta, C.~Lan, K.~Lee, B.~Lee, E.~Li, M.~Li, W.~Li, Y.~Li, J.~Li, H.~Lim, H.~Lin, Z.~Liu, F.~Liu, M.~Maggioni, A.~Mahendru, J.~Maynez, V.~Misra, M.~Moussalem, Z.~Nado, J.~Nham, E.~Ni, A.~Nystrom, A.~Parrish, M.~Pellat, M.~Polacek, A.~Polozov, R.~Pope, S.~Qiao, E.~Reif, B.~Richter,
  P.~Riley, A.~C. Ros, A.~Roy, B.~Saeta, R.~Samuel, R.~Shelby, A.~Slone, D.~Smilkov, D.~R. So, D.~Sohn, S.~Tokumine, D.~Valter, V.~Vasudevan, K.~Vodrahalli, X.~Wang, P.~Wang, Z.~Wang, T.~Wang, J.~Wieting, Y.~Wu, K.~Xu, Y.~Xu, L.~Xue, P.~Yin, J.~Yu, Q.~Zhang, S.~Zheng, C.~Zheng, W.~Zhou, D.~Zhou, S.~Petrov, and Y.~Wu, ``Palm 2 technical report,'' 2023.

\bibitem{anthropic2023claude3}
{Anthropic}, ``Claude 3 model card,'' tech. rep., Anthropic, 2023.

\bibitem{openai2024gpt4technicalreport}
OpenAI, J.~Achiam, S.~Adler, S.~Agarwal, L.~Ahmad, I.~Akkaya, F.~L. Aleman, D.~Almeida, J.~Altenschmidt, S.~Altman, S.~Anadkat, R.~Avila, I.~Babuschkin, S.~Balaji, V.~Balcom, P.~Baltescu, H.~Bao, M.~Bavarian, J.~Belgum, I.~Bello, J.~Berdine, G.~Bernadett-Shapiro, C.~Berner, L.~Bogdonoff, O.~Boiko, M.~Boyd, A.-L. Brakman, G.~Brockman, T.~Brooks, M.~Brundage, K.~Button, T.~Cai, R.~Campbell, A.~Cann, B.~Carey, C.~Carlson, R.~Carmichael, B.~Chan, C.~Chang, F.~Chantzis, D.~Chen, S.~Chen, R.~Chen, J.~Chen, M.~Chen, B.~Chess, C.~Cho, C.~Chu, H.~W. Chung, D.~Cummings, J.~Currier, Y.~Dai, C.~Decareaux, T.~Degry, N.~Deutsch, D.~Deville, A.~Dhar, D.~Dohan, S.~Dowling, S.~Dunning, A.~Ecoffet, A.~Eleti, T.~Eloundou, D.~Farhi, L.~Fedus, N.~Felix, S.~P. Fishman, J.~Forte, I.~Fulford, L.~Gao, E.~Georges, C.~Gibson, V.~Goel, T.~Gogineni, G.~Goh, R.~Gontijo-Lopes, J.~Gordon, M.~Grafstein, S.~Gray, R.~Greene, J.~Gross, S.~S. Gu, Y.~Guo, C.~Hallacy, J.~Han, J.~Harris, Y.~He, M.~Heaton, J.~Heidecke, C.~Hesse, A.~Hickey,
  W.~Hickey, P.~Hoeschele, B.~Houghton, K.~Hsu, S.~Hu, X.~Hu, J.~Huizinga, S.~Jain, S.~Jain, J.~Jang, A.~Jiang, R.~Jiang, H.~Jin, D.~Jin, S.~Jomoto, B.~Jonn, H.~Jun, T.~Kaftan, Łukasz Kaiser, A.~Kamali, I.~Kanitscheider, N.~S. Keskar, T.~Khan, L.~Kilpatrick, J.~W. Kim, C.~Kim, Y.~Kim, J.~H. Kirchner, J.~Kiros, M.~Knight, D.~Kokotajlo, Łukasz Kondraciuk, A.~Kondrich, A.~Konstantinidis, K.~Kosic, G.~Krueger, V.~Kuo, M.~Lampe, I.~Lan, T.~Lee, J.~Leike, J.~Leung, D.~Levy, C.~M. Li, R.~Lim, M.~Lin, S.~Lin, M.~Litwin, T.~Lopez, R.~Lowe, P.~Lue, A.~Makanju, K.~Malfacini, S.~Manning, T.~Markov, Y.~Markovski, B.~Martin, K.~Mayer, A.~Mayne, B.~McGrew, S.~M. McKinney, C.~McLeavey, P.~McMillan, J.~McNeil, D.~Medina, A.~Mehta, J.~Menick, L.~Metz, A.~Mishchenko, P.~Mishkin, V.~Monaco, E.~Morikawa, D.~Mossing, T.~Mu, M.~Murati, O.~Murk, D.~Mély, A.~Nair, R.~Nakano, R.~Nayak, A.~Neelakantan, R.~Ngo, H.~Noh, L.~Ouyang, C.~O'Keefe, J.~Pachocki, A.~Paino, J.~Palermo, A.~Pantuliano, G.~Parascandolo, J.~Parish, E.~Parparita,
  A.~Passos, M.~Pavlov, A.~Peng, A.~Perelman, F.~de~Avila Belbute~Peres, M.~Petrov, H.~P. de~Oliveira~Pinto, Michael, Pokorny, M.~Pokrass, V.~H. Pong, T.~Powell, A.~Power, B.~Power, E.~Proehl, R.~Puri, A.~Radford, J.~Rae, A.~Ramesh, C.~Raymond, F.~Real, K.~Rimbach, C.~Ross, B.~Rotsted, H.~Roussez, N.~Ryder, M.~Saltarelli, T.~Sanders, S.~Santurkar, G.~Sastry, H.~Schmidt, D.~Schnurr, J.~Schulman, D.~Selsam, K.~Sheppard, T.~Sherbakov, J.~Shieh, S.~Shoker, P.~Shyam, S.~Sidor, E.~Sigler, M.~Simens, J.~Sitkin, K.~Slama, I.~Sohl, B.~Sokolowsky, Y.~Song, N.~Staudacher, F.~P. Such, N.~Summers, I.~Sutskever, J.~Tang, N.~Tezak, M.~B. Thompson, P.~Tillet, A.~Tootoonchian, E.~Tseng, P.~Tuggle, N.~Turley, J.~Tworek, J.~F.~C. Uribe, A.~Vallone, A.~Vijayvergiya, C.~Voss, C.~Wainwright, J.~J. Wang, A.~Wang, B.~Wang, J.~Ward, J.~Wei, C.~Weinmann, A.~Welihinda, P.~Welinder, J.~Weng, L.~Weng, M.~Wiethoff, D.~Willner, C.~Winter, S.~Wolrich, H.~Wong, L.~Workman, S.~Wu, J.~Wu, M.~Wu, K.~Xiao, T.~Xu, S.~Yoo, K.~Yu, Q.~Yuan,
  W.~Zaremba, R.~Zellers, C.~Zhang, M.~Zhang, S.~Zhao, T.~Zheng, J.~Zhuang, W.~Zhuk, and B.~Zoph, ``Gpt-4 technical report,'' 2024.

\bibitem{khot2018scitail}
T.~Khot, A.~Sabharwal, and P.~Clark, ``Scitail: A textual entailment dataset from science question answering,'' in {\em Proceedings of the AAAI Conference on Artificial Intelligence}, vol.~32, 2018.

\bibitem{liu2019multitask}
P.~Liu, X.~Qiu, and X.~Huang, ``Multi-task deep neural networks for natural language understanding,'' in {\em Proceedings of the 57th Annual Meeting of the Association for Computational Linguistics}, pp.~4487--4496, 2019.

\bibitem{williams2020anli}
A.~Williams, N.~Nangia, and S.~R. Bowman, ``Anli: A new benchmark for natural language understanding,'' in {\em arXiv preprint arXiv:2004.07828}, 2020.

\bibitem{camburu2018snli}
O.-M. Camburu, T.~Rockt{\"a}schel, T.~Lukasiewicz, and P.~Blunsom, ``e-snli: Natural language inference with natural language explanations,'' in {\em Advances in Neural Information Processing Systems}, pp.~9539--9549, 2018.

\bibitem{chung2022scaling}
H.~W. Chung, L.~Hou, S.~Longpre, B.~Zoph, Y.~Tay, W.~Fedus, Y.~Li, X.~Wang, M.~Dehghani, S.~Brahma, A.~Webson, S.~S. Gu, Z.~Dai, M.~Suzgun, X.~Chen, A.~Chowdhery, A.~Castro-Ros, M.~Pellat, K.~Robinson, D.~Valter, S.~Narang, G.~Mishra, A.~Yu, V.~Zhao, Y.~Huang, A.~Dai, H.~Yu, S.~Petrov, E.~H. Chi, J.~Dean, J.~Devlin, A.~Roberts, D.~Zhou, Q.~V. Le, and J.~Wei, ``Scaling instruction-finetuned language models,'' 2022.

\bibitem{camburu2018esnlinaturallanguageinference}
O.-M. Camburu, T.~Rocktäschel, T.~Lukasiewicz, and P.~Blunsom, ``e-snli: Natural language inference with natural language explanations,'' 2018.

\end{thebibliography}
%%% and comment out the ``thebibliography'' section.

%%% Comment out this section when you \bibliography{references} is enabled.

\end{document}